# Document Automation Architectures and Technologies: A Survey

Mohammad Ahmadi Achachlouei[1*], Omkar Patil[*], Tarun Joshi, Vijayan N. Nair

Corporate Model Risk, Wells Fargo, USA

**Abstract**

This paper surveys the current state of the art in document automation (DA). The objective of DA is to reduce the manual effort during the generation of documents by automatically integrating input from different sources and assembling documents conforming to defined templates. There have been reviews of commercial solutions of DA, particularly in the legal domain, but to date there has been no comprehensive review of the academic research on DA architectures and technologies. The current survey of DA reviews the academic literature and provides a clearer definition and characterization of DA and its features, identifies state-of-the-art DA architectures and technologies in academic research, and provides ideas that can lead to new research opportunities within the DA field in light of recent advances in artificial intelligence and deep neural networks.

## 1 INTRODUCTION

Documents capture evidence and knowledge and convey information necessary for successful business processes that create value for stakeholders (Glushko & McGrath 2008). However, the manual steps involved in the creation of documents can be very time-consuming, resource-intensive, and prone to human error. Documents such as legal, technical, and clinical reports are usually highly structured and standardized, opening up a large potential to automate them.

Document automation (DA) aims to reduce this manual effort in the document generation process by automatically integrating input from different sources and assembling documents conforming to pre-defined templates. Automating the production of documents can save time, reduce cost, minimize risks, improve document quality, and minimize the human error that repetitive typing might introduce.

There have been reviews of commercial solutions of DA, particularly in the legal domain (Glaser et al. 2020; Dale 2019; Dale 2020), but to date there has been no comprehensive review of the academic research on DA architectures and technologies.

The current survey of DA aims to achieve the following objectives:

- Provide a clearer definition and characterization of DA and its features
- Identify state-of-the-art DA architectures and technologies in academic research
- Specify existing issues and bottlenecks in DA systems that can lead to new research opportunities

---

[1] E-mail address: mo.achachlouei@wellsfargo.com
[*] First and second authors contributed equally.

To conduct the current survey, we used Google Scholar to target the peer-reviewed journals and conference proceedings published by Elsevier, IEEE, ACM, Springer, and Taylor & Francis. The keywords used in the search comprise combinations of document automation, document assembly, document generation, document engineering, report generation, template construction, and natural language generation (NLG). Our main focus was on template-based automation of structured documents (e.g. legal, technical and scientific documents), but we also reviewed prominent studies on non-template NLG approaches which utilize linguistic grammar or end-to-end neural methods for data-to-text generation systems. We identified nearly 500 papers and, after reviewing paper abstracts, selected about 250 papers for further review. At the end, we selected about 50 papers with highest relevance to our survey goals.

The survey is organized as follows. Section 2 provides an overview of document definitions and standards as well as definitions of DA and its features. Section 3 summarizes the DA architectures and technologies identified in the reviewed studies. Section 4 discusses key findings, emerging trends, similarities, and differences of the reviewed DA architectures, as well as the relevance of each architecture for different applications, and proposes future research directions.

## 2 DEFINITIONS AND STANDARDS

### 2.1 DOCUMENT DEFINITIONS

**Document** refers to "a writing conveying information" (Merriam-Webster 2021) that has content, structure, and presentational characteristics. We can identify a spectrum of document types (Glushko & McGrath 2008) with highly narrative style documents (e.g. novels) at one end and highly transactional documents (e.g. purchase receipts) at the other end and a variety of hybrid documents in between, including semi-narrative and semi-transactional documents such as legal documents and technical/scientific reports.

The document engineering community defines a document as the union of two components: content and presentation (Gomez et al. 2014). The **document content** includes a template that defines the logical structure of the document, plus the components that instantiate the template. The **document presentation** includes the layout that defines exactly where each piece of content is to be placed and also how the piece will appear in the document.

**Automation** refers to the "automatically controlled operation of an apparatus, process, or system by mechanical or electronic devices that take the place of human labor" (Merriam-Webster 2021)

**Document automation** is not a well-understood, clearly defined term (Lauritsen 2012). Document automation is the term used by many commercial tools that aim to automate legal documents (for a survey of such tools, see Glaser et al. 2020). Table 1 provides several definitions from previous studies. Document automation usually refers to the automation of complex document process and workflow including template development and document assembly, usually through the use of advanced technologies.

*Table 1. A selection of document automation definitions*

| Study | Definition of document automation |
|---|---|
| Lauritsen 2007 | Document automation suffers from overbreadth. It sometimes refers to complementary technologies such as document management, comparison, and analysis tools, and word processing features like automatic numbering and cross-references. And it again implies a degree of automaticity that doesn't comport with the mixed initiative nature of present day and emerging drafting technologies. |
| Colineau et al. 2013 | There are commercial document automation systems (e.g., HotDocs, Exari, and Arbortext) which have been used in the legal profession to automate the production of custom-built legal documents (e.g., deeds of sale, standardized agreements, etc) and in the technical documentation field to produce model-specific product documentation. These tools provide mail-merge-like features extended with conditional inclusion/exclusion of coarse-grained text units generally on the order of sections, paragraphs or perhaps sentences.<br><br>DA systems provide tools for two tasks:<br>- **Template construction** – In this task, trained authors construct reusable templates, often from existing documents.<br>- **Document assembly** – In this task, an application uses an interview system to collect relevant characteristics from the user and then automatically constructs a personalized document for that user based on an appropriate template. |
| Dale 2018 | Document automation in the legal sector means generating routine legal documents. Document automation systems typically use some kind of fill-in-the-blanks templating mechanism that enables the creation of a legal document tailored to specific criteria. In some cases, the data required to generate the document is obtained via an iterative question-and-answer dialog. |
| Glaser et al. 2020 | The term Document Automation describes the trend of applying software solutions to automate the generation of documents. This study focuses at document generation as a sub-process of document automation. In most cases legal documents are highly structured and the decision which paragraphs are included depends on strict rules that have been defined in advance. The underlying logic for the document structure is usually modified only when the law or regulations change. For this reason there is a high potential to use document automation in the legal domain. |

## 2.2 DOCUMENT FORMATS AND STANDARDS

Digital documents can be represented, stored and exchanged in a variety of formats defined by standardization organizations such as W3C[2] and OASIS[3] (Hackos 2016). DA systems usually support standard formats such as PDF, TeX (Latex), DOCX, HTML, and IPYNB when converting documents from one format to another and generating final documents. DA systems may also utilize standard data-exchange formats such as XML (Extensible Markup Language) and JSON (JavaScript Object Notation) to represent documents in a way which is more machine-readable and more convenient for automated processes.

XML is a standard for defining markup languages with a set of start and end 'tags' which can be used to add more information about the main textual content, such as the mode of presentation or semantic information. The structure and allowable elements of an XML document can be defined in an XML DTD[4] file. XML structures can be mapped into HTML, plain text, or other XML structures using an Extensible Style Sheet Transformation (XSLT), which can be used to convert an XML document into other formats recursively. Table 2 lists selected XML-based standards highlighted in the reviewed studies.

The reviewed studies have also employed XML-based Semantic Web standards (Hitzler 2021), including the following W3C standards:

- RDF (Resource Description Framework) is a simple triple-based data model (triple of subject, predicate, object). RDF provides a graph-based formalism for representing metadata.

---

[2] W3C: World Wide Web Consortium
[3] OASIS: Organization for the Advancement of Structured Information Standards
[4] DTD: Document Type Definition

- OWL (Web Ontology Language) is a de-facto standard for ontology development. It provides a rich vocabulary to add semantics and context and allow reasoning and inference.
- SPARQL is an RDF query language able to retrieve and manipulate data stored in RDF.

*Table 2. Overview XML-based document standards*

| Standard | Description | Developed by | Extended from |
|---|---|---|---|
| OOXML | Office Open XML is a zipped, XML-based file format developed by Microsoft for representing word processing documents (docx), spreadsheets (xslx), charts, presentations and. The format was initially standardized by Ecma, and by the ISO and IEC in later versions. (Wikipedia) | Microsoft, ECMA, ISO/IEC<br><br>Standard: ECMA-376, ISO/IEC 29500 | XML, DOC, WordProcessingML |
| DITA | The Darwin Information Typing Architecture specification defines a set of document types for authoring and organizing topic-oriented information, as well as a set of mechanisms for combining, extending, and constraining document types. | OASIS | XML, HTML |
| DocBook | DocBook is a semantic markup language for technical documentation. It was originally intended for writing technical documents related to computer hardware and software, but it can be used for any other sort of documentation. | OASIS | XML |
| ODF | The Open Document Format (ODF), also known as OpenDocument, is an XML-based open source file format for saving and exchanging text, spreadsheets, charts, and presentations. The ZIP-compressed XML files saved under ODF, termed "OpenDocuments," have easily recognizable extensions, similar to Microsoft's proprietary .doc or .xls. | OASIS<br><br>Standard: ISO/IEC 26300; (OASIS OpenDocument Format) | XML |

# 3 DA ARCHITECTURES AND TECHNOLOGIES

## 3.1 REFERENCE ARCHITECTURE FOR DA

In order to better understand the various DA architectures reviewed in the following sections, first we introduce a reference architecture in this section to present a common vocabulary for DA components and their relationships. Using this common vocabulary, it will be easier to identify and discuss what components and aspects are supported or excluded in a specific DA architecture reviewed in the current survey. Our proposed reference architecture is depicted in Figure 1.

*Document design:* A fundamental characteristic of DA architectures is the schema of the document to be automated, and the associated ontology (if any). The schema defines how the document structure should be and what 'tags' it can carry. Ontology gives a meaning to the tags and defines relationships between them. These both comprise of what is shown in the diagram as 'document design'.

*Template*: The templates used for document automation are built on top of the document schema. Templates encode any static information to be added to all the documents, and have placeholders for the variable content which can be filled by the users or from external data repositories. Templates are not exhaustive in specifying the default content of the document, as it is often done by the 'document conversion' part of the architecture.

*External data sources* refers to any data source, such as a database, set of rules or semantic data such as RDF triplets which is used as a part of the document automation workflow.

*Content processing* defines a layer or a wrapper over that data to process it, either to assimilate it into the pipeline or to generate inferences. Most architectures fill the template with the data acquired from external sources before presenting the draft document to the user, although there are several which present a form to the user based on the template and the external data to modify the end document in certain ways.

*Document assembly*: Finally when the user(s) provide input, 'document assembly' takes place where the different fragments of documents are assembled based on the configuration defined. The assembly could also involve data directly from external data sources.

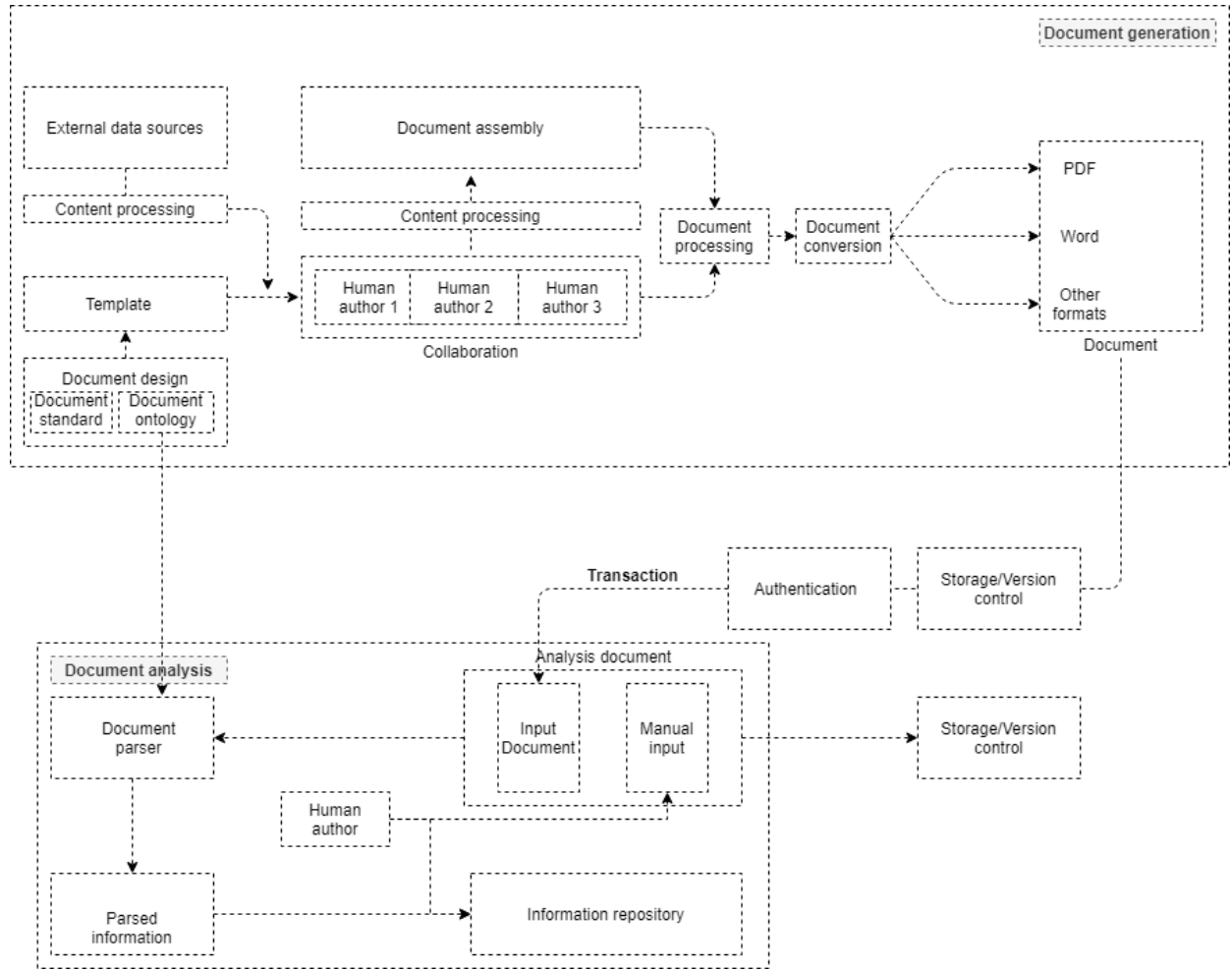

*Figure 1. Proposed Reference Architecture for Document Automation*

*Document processing*: After the 'document assembly' phase, the document is processed in the 'document processing' stage where citations, references, generating code outputs can be handled. Finally the document is sent to the 'document conversion' stage, where the draft is converted to the desired document output format.

*Document analysis*: In the 'document analysis' setting, the 'document parser' uses the defined document ontology to parse the input document into RDF triplets or some other convenient format. These triplets can be further processed to store in a repository or act on the data.

*Storage/version control* and *authentication* can be implemented as desired.

Human input in the analysis phase can happen either directly on the input document (if it is editable) or on the information extracted after parsing the document.

## 3.2 LEGAL DA ARCHITECTURES

In this section, we will review DA architectures aimed at document generation approaches used for creating legal documents.

### 3.2.1 Legal DA features

In this section, we will summarize legal DA features. Reviewing the history of legal DA evolution, Lauritsen (2012) identifies three main branches for legal DA:

**Document creation**: Technologies that assist with the creation of documents. There are two main areas of interest in document creation: (a) *Document authoring* (Power drafting), which includes word processing tools and a variety of other kinds of power tools that people can use to create documents. (b) *Document assembly* (auto assembly), including semi-autonomous document generation tools, e.g. HotDocs.

**Document analysis**: Technologies that assist in analyzing existing documents in various ways. There are three main areas of interest in document analysis: (a) *Disassembly*, which involves taking a document or a whole repository of documents and deconstructing it into its constituent parts for the sake of analysis and consequently for the purpose of guiding future document creation. (b) *Validation*, which includes a form of analysis which takes an existing document and validates it against some standard or some other document. (c) *Meaning extraction*, which goes beyond mere structural disassembly or deconstruction of a document and instead it takes a document that exists and parses it to extract and inference meaning out of it for the following purposes: What is the propositional content of the document? What rights and obligations are being expressed by the words? What are the parties agreeing to? What normative content is contained in the document?

**Document management**: this means managing documents once they already exist: (a) *Content storage and retrieval*, which includes managing the content itself, storing it, retrieving it, and allowing people to search. (b) *Rights and obligations*, which is related to managing the rights and obligations that are contained in legal documents (also called contract management software). How do we deal with them? How do we keep track of them and find them? Henley (2020) describes 16 criteria when evaluating legal DA platforms as shown in Table 3.

*Table 3. Overview of legal DA core and supporting features and evaluation criteria*

| Criteria for evaluation of legal DA platforms Henley (2020) | Legal DA core and supporting features (Glaser et al. 2020) |
|---|---|
| - Dynamic questionnaire* | *(a) Legal DA Core Features:* |
| - Capture and reuse inputs | - Template Structure |
| - Ability to pull information from databases | - Template Creation |
| - Ability to add help text in the questionnaire | - Document Generation |
| - Ability to generate multiple documents from one questionnaire | - Conditional Logic |
| - Questionnaire flexibility | - Formulas |
| - Ability to calculate results | - Documentation |
| - Ability to handle conditional logic | - Usability |
| - Ability to gather and process lists/repeats | *(b) Legal DA Supporting Features:* |
| - Ability to complete PDF forms | - Integration |
| - Ability to handle inserted templates | - Document Management |
| - Familiarity with template development environment | - User Management |
| - Cloud or desktop | - Collaboration |
| - Ability to gather data from others | - Enterprise Scalability |
| - Stability of the vendor | - Advanced Authentication |
| - Technical support and training | - Data Ownership |

*  Questionnaire here refers to any data input screen or interview involved in the assembly process.

Glaser et al. (2020), which evaluated 13 commercial legal DA tools, has specified two categories of legal DA features (Table 3): **Core DA features** are essential parts of the template creation process or document automation process. **DA supporting features** extend the core functionality of document automation tools and relate to the tool's integration into the existing landscape from a technical and an organizational point of view. More details about these core and supporting features can be found in Glaser et al. (2020).

Glaser et al. (2020), in their evaluation of 13 DA tools, identify two major ways that tool vendors have implemented templates: (The preference for one type or the other depends on the requirements in the individual case of organizations)

- The first approach uses Add-Ins for Microsoft Word to annotate existing DOCX files with custom tags that are used as placeholders or to mark the beginning and end of conditional text blocks. This approach relies on users' familiarity with Microsoft Word and delegates large parts of the template creation like formatting to the Word editor.
- The second approach is to utilize a custom editor that is built from scratch or templates that enforce a high degree of structure. This makes it easier to create templates that adhere to a strict structure but can also limit the number of structures and styles that can be modeled in the template. Templates are modularized to a varying degree with a higher modularization leading to more reusability between templates.

**A leading commercial legal DA tool**: Hotdocs, which appears in top Google Trends' queries related to DA, allows users to turn regularly used documents and forms into intelligent templates that guide the document creator through an interview, resulting in the production of an accurate, formatted document.

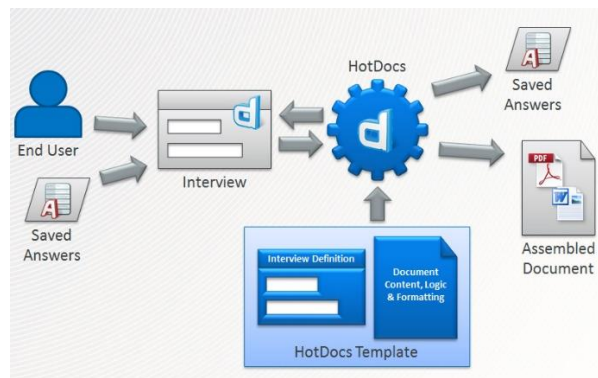

*Figure 2. An example legal DA workflow (HotDocs 2021)*

HotDocs Hub[5] is a modular suite of commonly-needed components for HotDocs integrations. Applications that integrate with HotDocs typically provide several additional features for controlling how HotDocs resources are used, such as HotDocs template storage, user authentication and authorization, and HotDocs interview generation. The Hub supplies these components without any additional development effort on the user's part.

### 3.2.2 Legal XML architectures (Akoma Ntoso)
Palmirani & Vitali (2011) define the pillars of the Akoma Ntoso architecture and also present basic elements of Akoma Ntoso XML standard. This standard can be used for marking up a legal document respecting a clear subdivision between the semantic layers of the knowledge embedded in it. Palmirani & Vitali (2011) define the following steps for legal analysis of the document:

- Identify the type of document
- Distinguish content from metadata and the presentation layer
- Identify the document's main legal components
- Define the document's URI
- Isolate each main legal block in the document
- Identify side notes by the author or issuing authority
- Detect and mark up the text's semantic elements

---

[5] https://help.hotdocs.com/hotdocshub/onpremise/earlier/1.5.0/help/admin/Understanding_the_HotDocs_Hub.htm

Marković & Gostojić (2020) produce legal documents in Akoma Ntoso format assembled using user-input and a set of legal rules. Palmirani and Governatori (2018) use Akoma Ntoso for marking up legal text, which then along with legal concepts and rules are used to check GDPR compliance for public sector cloud computing services.

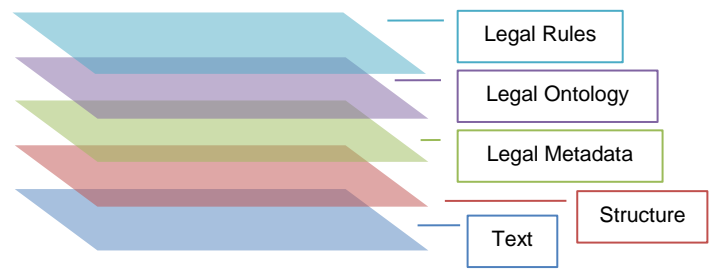

*Figure 3. Layers of representation in the Akoma Ntoso legal architecture (Palmirani & Vitali 2011)*

## 3.3 DITA: Darwin Information Typing Architecture

DITA is an XML-based architecture for creating documents based on maximum content re-use, multi-modal output and standardization. XML-based architectures promised three major enhancements for documentation pipelines (Priestley et al. 2001): (1) separation of form from content, (2) custom markup for organization-specific forms or features, and (3) a standard interface for creating and publishing documents using open-source tools. But often these claims are difficult to realize due to various reasons. For instance, the separation of form from content makes sure that the content once written can be published in various formats. But often, the content is structured according to a particular output form, and there is no guaranty that it will appear as expected in the other outputs too. The other two claimed advantages of using XML as a technology for documentation pipelines are in contrast to each other and result in tradeoffs. Essentially DITA considers these three advantages of using XML and seeks to fix them using certain principles so that they can be actually realized without any tradeoffs.

First, DITA advocates that the document authors forgo the conventional form of writing content structured as a book and instead write it in small chunks of information called 'topics'. The size of each topic could encompass a few paragraphs, but not as big as to cover a whole chapter in a book. Further, there could be three types of topics- 'tasks', 'concepts' or 'references'. For example a book could be published as a sequence of topics written by the author, while a slide presentation might choose only few of the essential topics for brevity. Expressing information as such topics creates a knowledge graph which allows the users multiple entry and exit points depending upon their interest in particular topic or the mode of presentation of the information.

Second, to optimize the creation of custom markup, DITA recommends a principle called 'specialization', which is similar to the 'inheritance' principle popular in object-oriented programming. This implies that all custom markup-up should inherit the basic markup from a parent and only define the changes in it. Creating a markup for a document essentially means creating a document definition that is a DTD (Document Type Definition) that will be followed by the authors while writing the content. This allows changes in the parent to be automatically replicated to all the child DTDs, which is often known as reuse by reference. To avoid the loss of standardization, the architecture limits you to redefine the markups already present in the parent DTD and not create an entirely new way of presenting information, as that would have no fallback for when shared outside the organization.

Third, to enhance the process followed for creation of end-documents from the XML-typed ones, DITA recommendations 'specialization-aware transforms' which advocates principles along similar lines of specialization. The stylesheets generalize to the most specific DTD it can transform in the hierarchy- so that organizations can render their content in a custom format by creating a specific transformation which inherits a general one.

Another architectural feature designed to make the process easier is that the DTDs and XSLTs are defined for each topic rather than the whole document. This allows us to break the process into components for re-use. DITA claims following these principles will not only ensure re-use of content and interchangeability, but also ensure the efficiency in creating custom markup and stylesheet transforms.

An application of DITA was presented by Eito-Brun (2020) where it was used to bring efficiency into software documentation in the Aerospace industry. DITA allowed efficient integration of information from various sources and single-source generation of HTML/PDF/Word documents. Data from various sources was collected and converted into XML using XSLT. XML files were then assembled and more content could be added using an XML editor. Final documents were generated using an XSLT stylesheet.

### 3.4 DPL: DOCUMENT PRODUCT LINE FOR VARIABLE CONTENT DOCUMENT GENERATION

Gomez et al. (2014) introduce DPLFW, a framework and tool supporting the Document Product Line (DPL) methodology for multi-user, variable content and reuse-based document generation. The goal is to provide a document generation alternative in which variants are specified at a high level of abstraction and content reuse can be maximized in high variability scenarios.

A DPL process starts with the development of a *document feature* model that defines the characteristics of a family of documents and the contributors involved (actors). This stage of the process called 'Document Engineering' entails a feature model defining content features and technology features as characteristics of the family of documents. Every content feature is linked to an 'info element' which is a reusable piece of content and eventually becomes a part of the document. Finally a document product line is set up which specifies how the info elements are integrated. Next in the 'Application Engineering' stage, the variability points are selected and the document product line is used to generate a document composed of the info elements. A specific document (an instance of the family) is created following a process in which the document components are taken from a repository, maximizing reuse. Penadés et al. (2014) defined a family of tax forms in DPL to customize the form according to the variability points selected by the user.

Figure 4 shows a proposed mapping of a general architecture for document product lines onto the reference DA architecture presented in Figure 1. The dotted lines and boxes in this particular architecture indicate that corresponding components (which are part of the reference DA architecture) are not explicitly present in the document product lines.

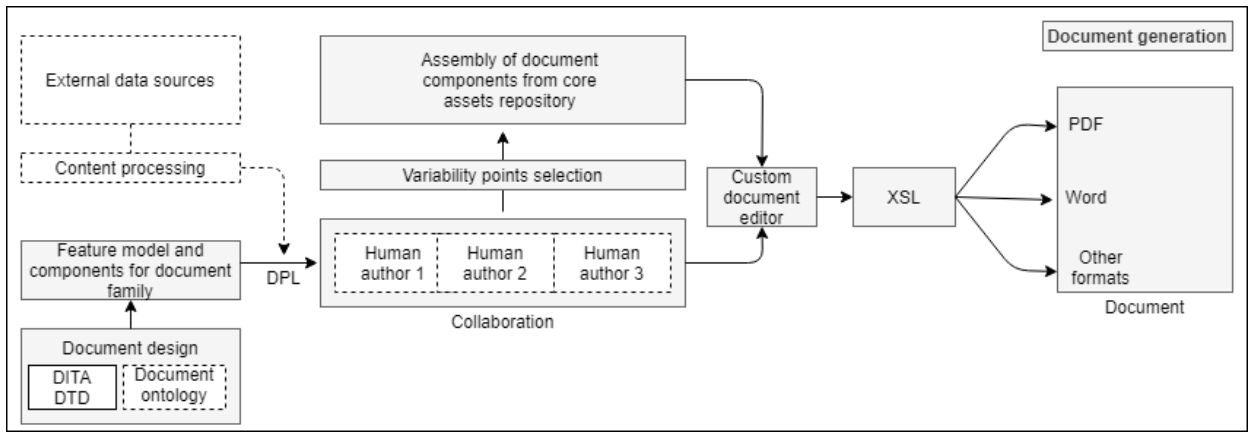

*Figure 4. Architecture for DPL (document product lines) – mapped to the reference DA architecture*

## 3.5 MODEL-BASED DOCUMENT GENERATION FOR SYSTEMS ENGINEERING

Delp et al. (2013) develop a workflow to generate documents from system models in NASA's model-based systems engineering projects which utilize the SysML modeling language. The workflow extends the SysML concepts of View and Viewpoint[6] in order to generate documents customized to the needs of the intended audience. The primary use case is to use the output of the viewpoint in DocBook to generate an HTML or PDF file according to the transformation applied. Delp et al. note that the technique can be used to integrate with existing or new analysis tools by adjusting the View format as an interface that can transmit subsets of model data back and forth with applications like Excel, Mathematica, Matlab, and more. This could further increase the collaboration between various teams.

Another example of a document automation solution built of top of MBSE setup in the Rubin observatory was given by Comoretto et al. (2020). A model based systems engineering (MBSE) approach is followed where the verification elements are defined in MagicDraw with the requirements as a starting point. All the test execution results are reported to Jira which are then propagated back to the verification elements and requirements. A tool called 'Docsteady' was developed to generate LaTeX test documents by extracting information from Jira using an REST API. Whenever changes are pushed to the document repository by the author, a CI service can be engaged to automatically render a PDF document and store it in 'LSST the Docs', a platform for hosting version controlled documents. The implemented documentation workflow presents three major advantages- reduced in time to produce V&V documents, better integration with the project's system engineering model and full traceability of system requirements.

Michot et al. (2018) proposed an architecture to keep traceability intact for bidirectional information transfer between document and the model. Development of APIs along with a tagging mechanism is used to synchronize corresponding elements between the model and the document.

In addition, there are other studies on model-to-document generation, such as Chammard et al. (2020). Figure 5 shows a proposed mapping of a general architecture for model-based document generation onto the reference DA architecture presented in Figure 1. The dotted lines and boxes in this particular architecture indicate that corresponding components (which are part of the reference DA architecture) are not explicitly present in the model-based document generation workflows.

---

[6] SysML offers two constructs for models: "Viewpoint" refers to the specification which a stakeholder provides for model elements and aspects that they are interested in. "Views" are how the stakeholders see the model according to the viewpoint specified.

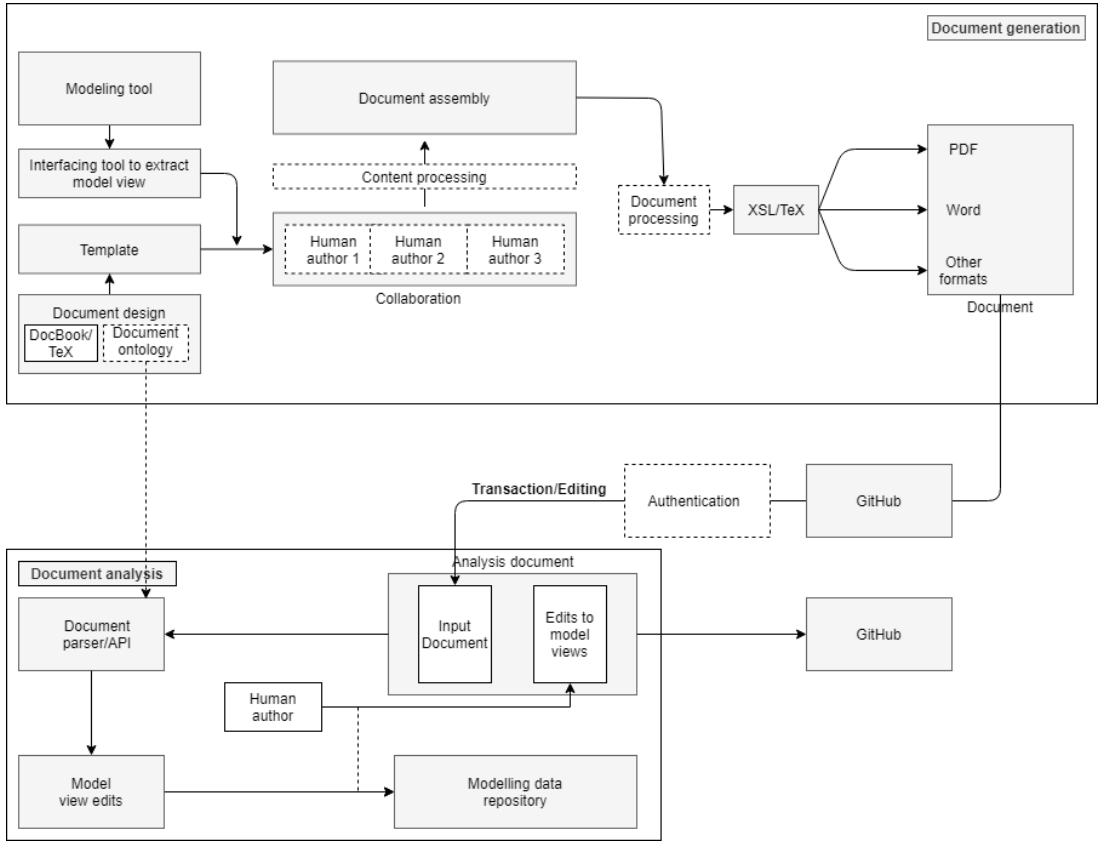

*Figure 5. Architecture for model-based systems engineering documentation workflows – mapped to the reference DA architecture*

## 3.6 KNOWLEDGE-BASED AND SEMANTIC WEB ARCHITECTURES

One category of DA architectures utilize knowledge repositories and reasoning mechanisms to derive inferences from a set of rules. The set of rules can be used together with user input, where users can make choices among the options presented to them. Such choices can impact the overall document assembly process. The inferences derived from the rules can have implications on how the document is assembled or what text is to be added to the document.

Marković & Gostojić (2020) propose a method for **knowledge-based document assembly**. They demonstrate this method by implementing a proof of concept with service contracts as a sample document type. Compared to other knowledge-based approaches, the method proposed by Marković & Gostojić (2020) uses an explicit formulation of legal norms prescribing the content and the form of service contracts to facilitate the document assembly process. The document assembly process includes two phases of analysis and synthesis. In the **analysis phase**, the legal professional and knowledge engineer formally represent a legal rule base and legal document templates to transform them into interview questions semi-automatically. The configuration of the document assembly process is also established in the analysis phase. The stated knowledge is represented as a set of rules in the LegalRuleML[7] format because of its relevance for the legal domain. The collected tacit knowledge is represented through ToXgene document templates that produce documents in the Akoma Ntoso format. The ToXgene format (Barbosa et al. 2002) is selected because it generates XML documents using a syntax similar to the industry-standard XML Schema. Akoma Ntoso is chosen for its flexibility in supporting multiple document types from a variety of legal systems. In the **synthesis phase**, the user answers the developed questions, and the system generates legal documents using the knowledge base and assembly configuration prepared from the analysis phase for the specific case. The system also generates an argument graph that explains how the document elements are derived from the legal rules and facts (as shown in Figure 6). The user answers needed by inference engine are stored in the RDF triples format, and those needed by the template engine are stored in a custom XML format as name-value pairs.

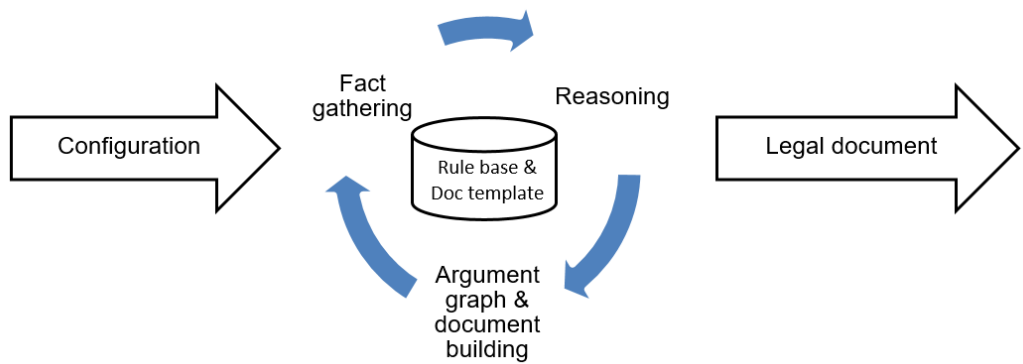

*Figure 6. The flow of knowledge-based document assembly process (Marković & Gostojić 2020)*

**Ontologies** find their place in document automation especially as enablers of enterprise-wide document semantic interoperability. Semantic interoperability is the process of representing, editing, and transmitting semantic information in one context, and then receiving and interpreting it in another. Several document ontologies have been proposed in the literature, such as DoCO by Constantin et al. (2016). DoCO is a document ontology model consisting of structural (container, block, inline etc.), rhetoric (caption, reference etc.), and hybrid (sentence, footnote, paragraph etc.) classes.

Yang et al. (2020) proposed a **cross-context tabular document representation approach**, called Tabdoc, to represent heterogeneous semantic documents across domains in a consistent and interoperable way. This approach aims to address existing limitations in the research methods of semantic interoperability, including standardization, ontology modeling, and collaboration templates. Yang et al. (2020) implement the proposed method under Sign Description Framework (SDF) and develop a new editor for semantic document creation. The Tabdoc framework ensures semantic interoperability by deconstructing the problem of representing semantic documents into three levels: vocabulary, relationship and document level.

One such application of ontology was given by Mirza & Sah (2017), where they developed a system to automatically check the format and structure of ACM SIG documents. The new Word format uses OOXML as the document standard, where the relevant metadata from document.xml and styles.xml is identified and converted into RDF (N3) triplets using an ontology developed for the conference documents. The ontology consists of 9 classes, 7 object properties and 67 data-type properties. The RDF

---

[7] LegalRuleML is an OASIS standard which defines a rule interchange language for the legal modeling and reasoning: https://www.oasis-open.org/committees/tc_home.php?wg_abbrev=legalruleml (Accessed 15.5.2021)

triplets consist of an index for document ID and part as the subject, the part feature extracted from the OOXML metadata as the predicate and the actual value extracted as the object. Once the RDF triplets are obtained, they are analyzed for developed rules using Jena reasoner- where additional triplets are added for the results of the validation by Jena. The number of rules was set according to the time it takes to execute them versus the validation requirements. Finally SPARQL is used to query these newly developed RDF result triplets and an XML document is generated. The authors validated the usefulness of the tool with a user study and noted that the format checking can take about 10 seconds, which is a considerable efficiency gain when compared to manual process.

Ontologies have also found application in integrating data models and information with document automation pipelines.

One such use case is for personalization of documents where Colineau et al. (2013) present a public administration information delivery system in Australia's Department of Human Services (DHS) that produces websites that are tailored to individual users. This system, called DHS-Myriad, includes a knowledge base with reasoning capabilities, a document planner and an authoring tool. The architecture consists of 4 major components- a query tool, document planner, context model and an authoring tool. The context model encodes the semantic information involved in the communication- the concepts and individuals from the domain (intended content), user, and possibly the discourse history and computation device. First a query form is presented to the user where the questions are created from the context model using an ontology verbalizer called the Semantic Web Authoring Tool (SWAT), developed by Third et al. (2011). The document planner creates a discourse tree structured with rhetorical relations using a text planner and declarative plan operators. Based on the choices made by the user in the query form, a tailored text is generated using the document planner. Finally an authoring tool is created to aid technical writers not comfortable with creating ontologies. The output of the system is a dynamic website tailored for a particular user. Referring to the system's usability test results, the authors note that the output of the tailored delivery system has been understandable and useful to users.

Another more direct use of ontologies for integration with data sources is provided by Pikus et al. (2019), where they develop a semi-automatic end-to-end documentation system, able to generate documents based on structured data represented in RDF format. The system extracts distributed lifecycle data, for storing it in a concise semantic manner and for providing relevant information in topic-centric and variant-specific documents. As a use case for document generation, they employ the RDF-based lifecycle tool integration standard OSLC (Open Services for Lifecycle Collaboration), add extended publishing information and leverage DITA in order to generate target documents. In particular, Pikus et al. 2019 extend OSLC domain ontologies (RDF) with documentation-relevant triples and propose additional DITA elements for referencing RDF data. Only a subset of RDF objects are relevant for documentation – called PIOs (Publication Information Object). PIOs are referenced in DITA templates being bound to particular RDF classes. Finally, they formalize a PIO Management Environment (called PIOME) and propose a DITA template processing algorithm.

Figure 5Figure 7 shows a proposed mapping of a general architecture for knowledge-based and semantic document generation onto the reference DA architecture presented in Figure 1. The dotted lines and boxes in this particular architecture indicate that corresponding components (which are part of the reference DA architecture) are not explicitly present in knowledge-based and semantic document generation.

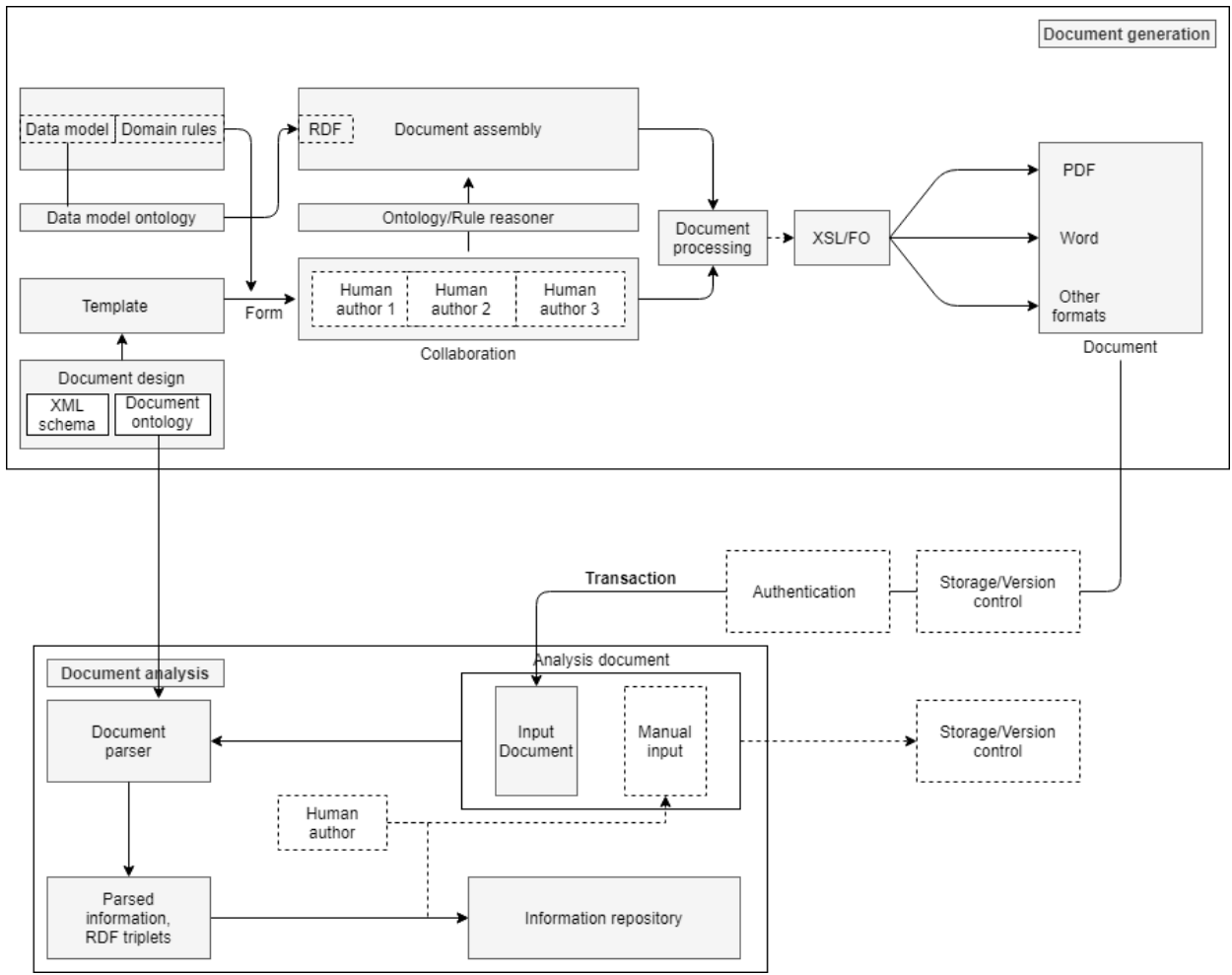

*Figure 7. Architecture for knowledge-based and semantic document generation – mapped to the reference DA architecture*

## 3.7 ARDEN SYNTAX FOR CLINICAL DOCUMENTS

**Arden syntax** is a domain-specific programming language designed to meet the requirements of clinical event-monitoring in the form of Medical Logic Modules (MLM). An **MLM** is a production rule with a frame-like structure made of categories and slots, used to make a single clinical decision. Arden syntax and the accompanying MLMs were designed for the identification and notification of clinical events. (Hripcsak 1994).

A number of studies have utilized and extended the Arden syntax to generate clinical reports. Kraus et al. (2016) create an intensive care discharge letter system by extending the current Arden syntax to include 'interface calls' for easy creation of templates. The Arden syntax can interact with the local patient data management system (PDMS) to fetch patient information from the electronic medical records. This information is used to fill in the template while generating a report. The extensions also allows intensivists to add more detail into their discharge reports such as outliers in sensor data and the presence of specific microorganisms, as shown in an example in the paper.

Arden syntax, which was originally created for clinical support decision systems, lacks necessary features needed for clinical document generation. For example, there was a need by intensivists to generate more sophisticated reports than what they could do via legacy simple templates. Kraus (2018) and Kraus et al.

(2019) extended Arden syntax to create a more general language suitable for all clinical applications called PLAIN (Programming Language, Arden-Inspired). They also created a PLAIN data markup language (PDML) as a convenient representation of the electronic medical records data and demonstrated the PLAIN and PDML in a report generation example. The advantages of PLAIN include:

- PLAIN provides a standard language to be used in the medical domain.
- It helps treat MLMs as user-defined functions (UDFs), and provides native support for RESTful web services (Fielding 2000).
- It introduces string interpolation, making it more convenient to create document templates over the existing methods of string manipulation in the Arden syntax.
- Future development of this extension would involve natural language generation for texts of very large size and offer more flexibility in sentence creation.

The Arden syntax is not based on XML, but it can be modeled in XML. For example, see Kim et al. 2008.

## 3.8 QUANTITATIVE ANALYSIS REPORTS

In this section, we will review DA architectures aimed at document generation approaches used for quantitative analysis reports, which are common in scientific studies and financial model risk management processes. This approach emphasizes the reproducibility of the analysis presented in the report, along with helping the author to weave the text with the code so that the analysis becomes presentable. Such architectures are expected to support the following features:

- Authoring tool to write text; add tables, figures and mathematical equations; and cross reference
- Ability to connect to databases to automatically add model meta-data and model dependency information
- Ability to add code snippets whose outputs need to be directly integrated into the document (literate programming)
- Ability to publish in PDF, Docx and HTML
- Collaboration features for commenting and track changes
- Version control and reproducibility

Such documentation approaches are usually created on top of the analysis tool or programming language that is primarily used by analyst document authors. For example, Lang (2001) present a way to embed a statistical programming language (called S) into an XSLT translator so that an XML document containing the code in that language can be converted into an interactive HTML document along with the code output. Following that Leisch (2002) presented a new report generation method called Sweave which permitted weaving statistical programming languages (S/R) into Latex so that the documentation for data analysis can be carried out seamlessly. Sweave automatically generated a complete Latex document by evaluating the statistical programming code (in S/R), and also generating a code preview for it, keeping the rest of the Latex code as it is. Another such software is Codebraid (Poore 2019) which executes code blocks and inline code in Pandoc Markdown documents (MacFarlane 2012) as part of the document build process. A single document can involve multiple programming languages, as well as multiple independent processes per language, including Jupyter kernels. Since Codebraid only uses standard Pandoc Markdown syntax, Pandoc handles all Markdown parsing and format conversions. There is also support for programmatically copying code or output to other parts of a document. Kane (2020) refers to literate programming as having both narrative and computational components, which is important to express the findings and research to a wider audience and also to provide more credibility to your work. To this end, they discuss an R package called 'listdown' which automatically creates R Markdown documents from a named list which contains data-structures including tables and graphs to be presented in hierarchical order.

### 3.8.1 Notebook-centric document automation

Architectures and technologies based on Jupyter notebooks are commonly utilized in scientific and quantitative analysis documentation workflows. Jupyter notebooks provide an interactive web application allowing users to create and share programmatic analysis in over 40 languages and provides features to add data commentary in the same environment (Kluyver et al. 2016; Pérez & Granger 2007). Jupyter notebooks contain live code, equations, visualizations, and text. A notebook supports a variety of quantitative analysis requirements, including data cleaning and transformation, statistical modeling, data visualization and documentation. Each notebook consists of several cells. Users can change the cell type of any cell in a notebook using the toolbar. The default cell type is Code. Users can write and execute analysis code in code cells and see the interactive output (including tables, plots, equations, etc.) below each cell. This interactive experience enhances usability and resembles the step-by-step nature of analysis. Users can also change a cell type to Markdown in which they can write text which can be formatted using the Markdown language.

Technically, a Jupyter notebook file (IPYNB) is a document format built on top of JSON. Each notebook communicates with a kernel process – which executes code – using a protocol called Interactive Computing Protocol.

Given the capabilities of Jupyter notebooks to support analysis and documentation in a single file, they can serve as standalone documents for scientific and quantitative analysis documentation to share data preparation procedures, analysis code, visual/tabular results and discussion in a single file. For example, BioJupies (Torre et al. 2018) is an online tool which creates Jupyter notebooks complete with narrative text, interactive visualizations and analysis for RNA sequence data. The tool can be used by novice users who input a raw data file (FASTQ), which is processed and a Jupyter notebook is created with the required analysis, along with storage and deployment options. Ragan-Kelly et al. (2013) show how notebooks were hosted on cloud to enable collaborative and interactive research on compute-intensive subjects such as genomics.

The open-source ecosystem has driven the development of numerous resources for notebook management and sharing, which make Jupyter Notebook a powerful analysis documentation tool. JupyterHub, for instance, facilitates centralized deployment and authentication for coordinated use of notebooks by a group of analysts (Kluyver et al. 2016). For storage and viewing, Nbviewer[8] facilitates sharing of notebooks in a read-only mode, while Binder[9] turns a GitHub repository of notebooks into a collection of interactive notebooks. Kluyver et al. (2016) discuss how for a scientific paper on detection of gravitational waves, a notebook was utilized to filter and process data and was made accessible to the public using Binder. From a computational perspective, Jupyter notebooks offer benefits to end-users as notebooks communicate with the kernel through a network protocol, implying that the compute power could be centrally managed for all end-users (Perkel 2018). Juneau et al. (2021) use a Jupyter-based architecture to conduct analysis in the Astro Data lab of NSF's National Optical-Infrared Astronomy Research Laboratory (NOIRLab). Typically in such a workflow, after user authentication, external data is queried and retrieved directly in the notebook, post which the analysis is carried out. Henderson et al. (2019) develop a system to avail high performance computing to end-users of high volume experimental science laboratories through Jupyter Notebooks. They propose an architecture based on JupyterHub service for utilizing HPC resources at the National Energy Research Scientific Computing Center (NERSC). Interfacing between Jupyter processes and compute nodes also allow for distributed computing using IPyParallel or Dask. Beg et al. (2021) attribute the effectiveness of Jupyter for reproducible research in computational sciences and mathematics to easily shareable nature, consolidation of code and narration and being able to execute them as a 'batch job' among other reasons. They also emphasize on the ease of creating documentation through Jupyter Notebooks with Sphinx[10], which creates an HTML file or PDF as a result. Beg et al. discuss a workflow for reproducible research based on Jupyter Notebooks, where once authored, the notebooks are stored in a repository such as GitHub, after which Binder is used to share that notebook with the public directly through the browser, without any additional installations. JupyterHub helps to manage the resources associated with Jupyter Notebook at an institutional level such as computation kernels along with adding a layer of authentication. Beg et al. state the drawbacks of Jupyter Notebooks as having an undefined cell execution order and high initial start-up effort and time. In an effort to understand and improve the reproducibility of Jupyter notebooks, Pimentel et al. (2021) design a JupyterLab extension called 'Julynter', which identifies potential issues and suggests modifications to improve the reproducibility of a notebook. They attribute most reproducibility failures to missing dependences, hidden states, incorrect execution order, and data accessibility.

---

[8] https://github.com/jupyter/nbviewer
[9] https://mybinder.org/

Jupyter Notebooks have also been considered as an intermediary form of documents and can be converted to a more conventional document format such as PDF or HTML. Several open source contributions have also taken place to that end to facilitate the conversion. While nbviewer creates static webpages with unresponsive widgets from notebooks, Nbinteract (Lau & Hug 2018) embeds JS code in a 'run widget' button in place of the static widget on the webpage. This JS code essentially creates a Jupyter Kernel on Binder (a free JN hosting service), and renders a live widget on the webpage. The service also handles all future communications between the widget and the Binder kernel. NbConvert[11] and IPyPublish (Sewell 2017), as shown in Figure 8, allow users to post-process the notebooks to generate a report in HTML, Latex and other formats using Jinja templating. The users add the analysis and prose in Jupyter Notebook which has extensive support for adding tables, images, equations and performing analysis. 'Ipywidgets' allows embedding widgets in the notebooks through which users can interact and modify the resulting document.

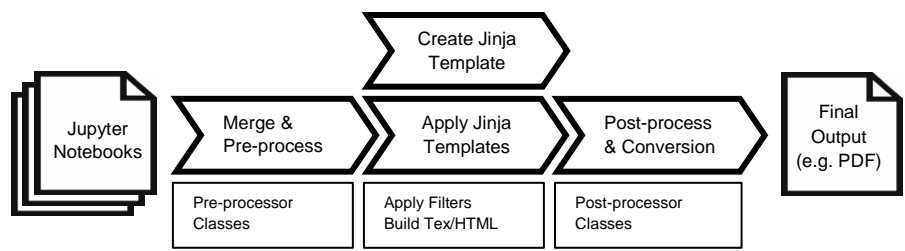

*Figure 8. IPyPublish notebook-centric architecture (Sewell 2017)*

### 3.8.2 Wells Fargo's Document Generation System (DGS)

Wells Fargo's Document Generation System (DGS) delivers document automation capabilities based on Jupyter Notebooks to help model validators create and edit model validation documents and publish the final PDF report. Using the DGS, users can automate repetitive and time-consuming tasks in the model validation process to enhance efficiency. Figure 9 shows building blocks of the DGS. For each model validation project, the DGS collects relevant data from model databases and provides a model-specific notebook template in which users can write their validation assessment with the possibility to integrate plots and tables from statistical analysis notebooks. In addition to common notebook extensions such as table of contents and spellchecker, the DGS offers custom notebook extensions (Figure 10) to provide a more advanced search and replace features and to enhance collaboration among multiple authors by tracking changes, merging documents, and adding comments. It also develops extensions to facilitate adding new cells tailored for adding figures, tables and equations. The main objectives and features of the DGS are as follows:

1. **Reproducibility and reusability**: To achieve this objective, the DGS utilizes version-controlled Jupyter notebooks for both analysis code and verbiage; it integrates the validation report notebooks with analysis notebooks; the reports can be reproduced later; users can easily adapt and reuse documents to generate validation reports for similar projects; the DGS provides a well-documented workflow as a starting point for new users.
2. **Efficiency improvement**: To achieve this objective, the DGS decouples presentation from content, integrates with the model database and other metadata sources, and offers possibilities to integrate with upstream model development processes and documents.
3. **Quality check and standardization**: To achieve this objective, the DGS offers automated standardization and quality checks for both content and presentation aspects of validation reports.

---

[10] https://www.sphinx-doc.org/en/master/index.html
[11] https://nbconvert.readthedocs.io/en/latest/

The DGS also utilizes the IPyPublish library (Sewell 2017) to convert Jupyter Notebooks into Latex/PDF and also HTML/DOCX using Jinja templates.

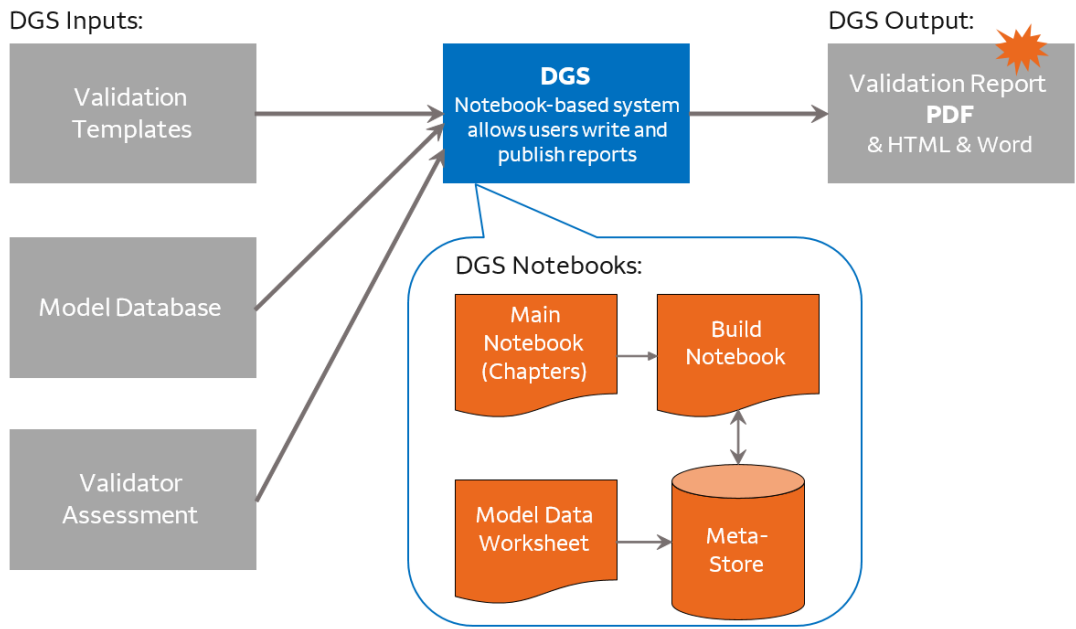

*Figure 9. DGS building blocks*

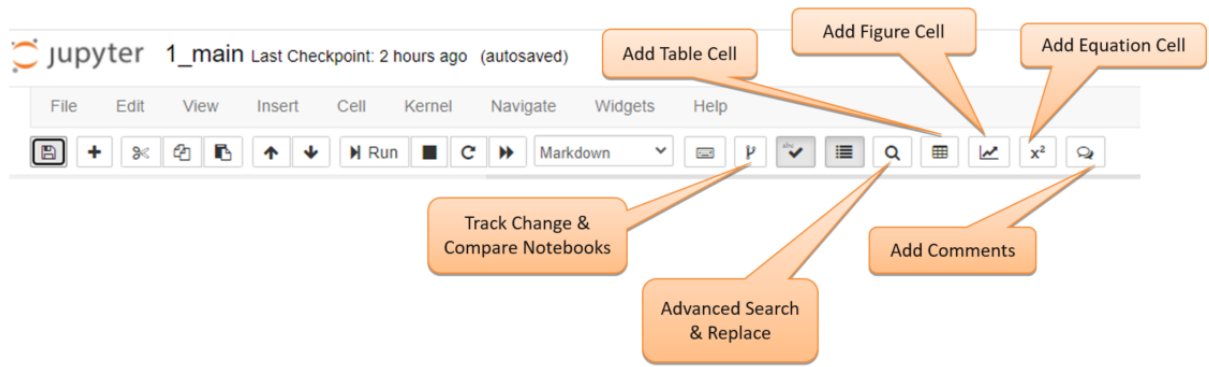

*Figure 10. DGS notebook extensions*

## 3.9 DATA-TO-TEXT NLG ARCHITECTURES

NLG can be used to generate periodic textual documents based on collected or measured data. In this section, we will summarize main building blocks of a standard NLG system, providing more details on realization approaches. We will also review several NLG architectures based on deep neural networks including language models. Then we will discuss the relationship between NLG and DA approaches and conclude with some applications of NLG for document automation

### 3.9.1 NLG building blocks

The classical NLG system architecture consists of three modules, which make up most NLG pipelines (Reiter & Dale 2000; Gatt & Krahmer 2018): **Content determination (or document planning)** analyzes the signals in the data and determine what messages to convey in the text. This can depend on what messages are considered interesting and on the type of the targeted audience. **Microplanning** involves choosing the particular words, syntactic constructs, and markup annotations used to communicate the information encoded in the document plan. Microplanning consists of text structuring, sentence aggregation, lexicalization and referring expression generation. Finally we have **surface realization** to generate the final text through grammatical, statistical or template methods.

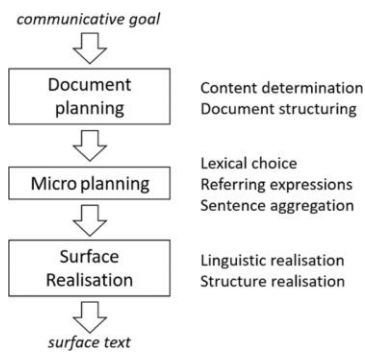

*Figure 11. NLG building blocks (adapted from Reiter & Dale 2000)*

### 3.9.2 NLG surface realization approaches

Gatt & Krahmer (2018) identify three categories of surface realization approaches:

1. *Template-based* approaches, which provide a boilerplate for the fixed parts of a document with variable slots that can be filled by data from user input and other data sources. These approaches are suitable when application domains are small and variation is expected to be minimal.
2. *Hand-coded grammar-based* approaches, which provide general-purpose, domain-independent realization systems based on a grammar of the language under consideration.
3. *Statistical* approaches, which acquire probabilistic grammars from large corpora, reducing the manual labor needed for hand-crafted grammars.

### 3.9.2.1 Template-based realization

Traditional template-based realizers can generate texts efficiently. But for some applications, templates are not flexible enough and are difficult to reuse (Reiter 1995). There are several studies on how to make traditional template-based realizers more powerful.

McRoy et al. (2003) developed a system called YAG (Yet Another Generator) based on augmented templates. YAG is a real-time, general purpose, template-based surface realizer that accepts underspecified inputs. YAG provides an expressive and declarative language for specifying templates, along with default template-based grammar for a core fragment of English. Each template in YAG consists of rules and slots, where a slot is filled with a value by the user or the application in real time, and the rules decide how to realize the surface representation, also supporting recursive and nested templates. Several rules exist such as the 'String Rule', which returns a pre-defined string as the result and the 'If rule', similar to the 'if' condition in many programming languages. YAG also provides the requisite speed to work with real time systems, as the generation time does not depend on the size of the grammar but on the complexity of the templates. Increasing the number of templates may give more expressive output through YAG, but also increase the execution time.

### 3.9.2.2 Grammar-based realization

Hand-coded grammar-based approaches, which provide general-purpose, domain-independent realization systems based on a grammar of the language under consideration. The rules of grammar govern both morphology (which determines how words are formed) and syntax (which determines how sentences are formed) (Reiter & Dale 2000). Table 4 provides a short summary of key grammar-based realization systems.

*Table 4. A summary of grammar-based realization systems*

| Realizer | Description |
| --- | --- |
| SimpleNLG (Gatt & Reiter 2009) | Java Library offering direct control over realization, such as over morphological operations and linearization. There is a clear separation between the syntactical and morphological operations, where the syntactical component covers the full range of English verbal forms and morphological rules are re-implemented from MORPHG (Minnen et al. 2001). |
| RealPro (Lavoie & Rambow 1997) | Implemented in C++ along with APIs in Java too, RealPro is based on meaning text theory (Kahane 1984). Meaning text theory has seven ordered levels of representation and language is modeled as a correspondence between meanings and sounds. RealPro converts the input 'Deep-Syntactic Structure' to a surface representation. |
| SURGE (Elhadad & Robin 1996) | SURGE is implemented in the special purpose programming language FuF and it is distributed as a package with a FuF interpreter. SURGE is based on functional unification grammar (Kay 1984). The 'functional unifier' in the interpreter uses syntactic features from the grammar to flesh out the input skeletal tree, which is then linearized following precedence constraints indicated in the tree. |
| KPML (Bateman, 1997) | KPML is a multilingual extension of the Penman system (Mann, 1983), based on systemic functional grammar (Halliday and Matthiessen 2013). Input to KPML can be provided in the form of sentence plans in SPL-Sentence plan Language (Kasper, 1989). |

### 3.9.2.3 Statistical surface realization

Lattice based methods put forward by Langkilde and Knight (1998) propose a two-step solution to sentence generation, where the first step is to create a lattice structure of different phrasal alternatives from a semantic input (PENMAN-style Sentence Plan Language (SPL) (Penman, 1989), with concepts drawn from the SENSUS ontology) to the system. Then a bigram (or trigram) model is used to select the right choice among the alternatives to come out with the best sentence. Each lattice structure represents all the possible sentences that can be created, and the sentence having the highest value for the joint probability of its words is chosen as the output, where the individual bigram probabilities are obtained from a corpus. All the possible inflections are left for the bi-gram model to filter and choose the right one. The conversion from AMR to word lattices happens using keyword based grammar rules, as opposed to grammar rules based on syntactical categories. This leaves the syntactical choice to the system and does not force the client to make the specifications. Langkilde (2000) further proposed an improvement to the lattice based filtering method by converting AMR into a forest representation, which yields huge improvement in the time-complexity of the algorithm without compromising the sentence quality. The forest-based filtering method has a bottom-up dynamic programming ranking algorithm for choosing the right words, where the crucial aspect is that the choice of words made for a sub-tree is the only thing of relevance and other alternatives can be discarded, significantly reducing the search space.

Several works followed that of Langkilde and Knight, such as from Bangalore and Rambow (2020) where they proposed FERGUS (Flexible Empiricist/Rationalist Generation Using Syntax). FERGUS uses a stochastic tree model to construct a lattice from input dependency tree, following which the best sentence is selected using a trigram language model. See Gatt & Krahmer (2018) for more developments in statistical realizers.

*NLG for generating neonatal care and weather reports*

Report generation using NLG becomes very useful in cases where the text is based on data which is generated periodically. Gatt et al. (2009) develop a system called BT-45 which generates patient data summaries spanning every 45 minutes for neonatal intensive unit care patients. Their architecture sketches out four blocks- signal analysis, data interpretation, document planning, and microplanning and realization. These blocks are supported by an ontology which captures much of the domain knowledge. The paper also explores creating different textual content for different stakeholders- doctors, nurses and parents/guardians. Another example of generating reports using NLG (using Arria NLG engine) was given by Sripada et al. (2014) where they generate weather reports for nearly 5000 locations for the UK national weather service- the Met office. The textual quality was found to be up to the mark after a two year assessment by a Met office expert and further user surveys.

### 3.9.2.4 Neural NLG: RNN-based architectures
With recent advances in deep learning methods, they have re-gained traction for NLG applications. Neural networks learn dense and low-dimensional. The ability to learn such representations automatically makes them well-suited to capture grammatical and semantic generalizations. Recurrent neural networks (RNNs) such as LSTMs are used for sequential modelling, where one application was for generation of grammatical English sentences using character-level LSTM by Sutskever et al. (2011). They introduce Multiplicative RNNs for generation of text using characters, which were chosen over morphemes for the ease of implementation. However, models that generate text from semantic or contextual inputs usually are of encoder-decoder architectures or are conditioned language models. Encoder-decoder architectures are well-suited to solve sequence-to-sequence (SEQ2SEQ) tasks such as machine translation, and Ferreira et al. (2017) adapted SEQ2SEQ models for generating text from abstract meaning representations (AMR). They first perform de-lexicalization, compression and linearization steps to generate flat AMR which is then realized using either a phrase-based machine translation (using Bayes rule) or neural machine translation having an RNN-based encoder-decoder architecture along with attention mechanism. Dušek &

Jurcicek (2016) use a SEQ2SEQ model with a re-ranker to produce two types of outputs from dialogue acts (Young et al., 2010).

Gatt and Krahmer (2018) noted that deep learning approaches have become popular for natural language generation, and dominant for tasks such as image captioning. The promise of stochastic and deep learning approaches relies on whether they can be scaled to work with large volumes of heterogeneous data and be used to generate long texts. Another question of importance is regarding the reusability of representations learnt during training for NLG applications, in comparison to image based applications where the features are reusable for a range of tasks.

*An example of neural NLG for DA: Automated auditing with machine learning*

Sifa et al. (2019) present a method to automate manual effort in ensuring completeness of financial documents according to legal requirements, for the financial auditing domain. Two documents are considered here: a requirements document, which lists down the parts that should necessarily be present in the document, and the actual document. Then the task boils down to finding the relevant paragraph in the document for each of the listed task, which is a manual effort performed by auditors. A machine learning based approach has been suggested to automate this very task to create a recommendation system which lists down the possible paragraphs for each of the requirements. This is done by first processing both the requirements and contents document semantically, that is by treating financial terms separately. Then a document representation is chosen, including n-grams, bag of words and neural language models. Unsupervised and supervised ranking approaches are proposed where the potential candidate paragraphs are ranked according to their likeness of being associated with a requirement. The system is also tested to incorporate the structural dependency between requirements and paragraphs.

### 3.9.2.5 Neural NLG: Pre-trained language models

*Why pre-trained language models?*

Despite the success of neural models for data to text generation and NLP tasks in general, they face several difficulties such as lack of large-sized task-specific datasets and the high computational cost required to train them (Vaswani et al. 2017). The transformer architecture was proposed by Vaswani et al. (2017) for parallelizable sequence to sequence generation. The transformer architecture constitutes of an encoder and a decoder both of which contain stacked self-attention and feedforward layers. Since then, several language models pre-trained on large corpuses of text have been developed based on the transformer model. Pre-trained language models are used for specific tasks often by fine-tuning the pre-trained models on the task-specific datasets (Devlin et al. 2018). These pre-trained language models produced state of the art result on several NLP tasks such as question answering, machine translation, text classification and abstractive summarization, as they obviate the requirement of large task-specific datasets, generalizing from unsupervised pre-training.

*Recent language models*

Radford et al. (2018) proposed a language model based on decoder part of the transformer architecture called Generative Pre-trained Transformer (GPT), where stack of multi-headed attention and feedforward layers produce a distribution over the output tokens. Subsequent versions of GPT (Radford et al. 2019; Brown et al. 2020) have grown in parameters and performance. Devlin et al. (2018) proposed BERT-'Bidirectional Encoder Representations from Transformers' which learns bidirectional context in a sentence to predict masked spans of text using the encoder part of the transformer. BERT also predicts a binary value for whether the second sentence in the input follows the first or not, helping it to learn sentence relations. Lewis et al. (2019) assert that BERT (Devlin et al. 2018) and GPT (Radford et al. 2018) are not optimal for the task of text generation as they predict missing spans of text independently and cannot learn bidirectional context respectively. They propose BART- 'Bidirectional and Auto-Regressive Transformers' which uses a bidirectional encoder and an auto-regressive decoder to generate text for tasks such as summarization and question-answering. Rather than first pre-training a model and then fine-tuning it separately on different tasks, Raffel et al. (2019) propose T5- 'Text-to-Text Transfer Transformer' to use the same model, along with the same loss function and hyper-parameters for all the tasks. Input to the T5 encodes a task-specific prefix, which allows the encoder-decoder transformer based model to treat all downstream tasks as text-to-text. Raffel et al. (2019) conclude that encoder-decoder architectures usually perform better than only-decoder architectures, and masking parts of the sentence to predict, worked best for pre-training.

*Language models for data-to-text NLG tasks*

For the purposes of this paper, we are concerned with converting structured input such as a graph or a table into text. Two relevant datasets for these tasks are ToTTo (Parikh et al. 2020) for generating text from highlighted cells in a table and WebNLG (Gardent et al. 2017) for generating text from RDF data. Li et al. (2021b) jointly train text generation from a knowledge graph using an encoder-decoder transformer based model and knowledge graph reconstruction using the decoder output to achieve state of the art result on WebNLG, followed by T5-Large (Raffel et al. 2019) in performance. Kale (2020) notes that T5 performs better than GPT-2 (Radford et al. 2019) and BERT on text generation tasks, and achieves state of the art for most if not all of them, including ToTTo using T5-3B. Mager et al. (2021) fine-tune GPT-2 to produce text from abstract meaning representation (AMR) by jointly training prediction of the text and reconstruction of AMR, the candidate sentences being rescored using cycle consistency on the AMR reconstructed. Overall, pre-trained language models have performed better than other methods for data–to-text generation and often produce near human quality output. For a survey on text-generation using pre-trained language models, see Li et al. (2021a).

### 3.9.3 Relationship between NLG and template-based DA

While NLG tools with linguistic knowledge (microplanning and grammar-based realization) can be utilized as a text generation component in DA systems, we can observe that most commercial NLG toolkits (similar to DA tools) utilize simpler NLG realization methods, i.e. templates to generate text.

Reviewing the commercial state-of-the-art of NLG in 2020, Dale (2020) notes that most of commercial data-to-text NLG products utilize similar mechanisms, which can be referred to as "*smart template*" mechanisms. For the kinds of commercial NLG applications available in 2020, much of the text in any given output can be predetermined and provided as boilerplate templates, with gaps to be filled dynamically based on per-record variations in the underlying data source. Comparing legal DA tools with commercial NLG tools, Dale (2020) believes that if we take a typical commercial NLG toolkit and add conditional inclusion of text components and maybe some kind of looping control construct, the resulting NLG toolkit becomes almost the same as the legal DA tools like HotDocs. Dale (2020) notes that linguistic knowledge and other refined ingredients of the NLG systems built in research laboratories, is sparse in commercial NLG tools and is generally limited to morphology for number agreement (e.g., one stock dropped in value vs. three stocks dropped in value).

There is an academic debate on template-based NLG approaches versus conventional NLG approaches (which employ more linguistic knowledge). Reiter (1995) highlights the benefits of conventional NLG approaches over template-based approaches. Van Deemter et al. (2005) challenge the common perception that template-based approaches to the generation of language are necessarily inferior to other NLG approaches as regards their maintainability, linguistic strength, and quality of output. Reiter (2016) instead of direct comparison between conventional NLG and template-based approaches, defines several levels of sophistication for generating text:

- *Level 1*: This level is closer to template-based approaches, including simple fill-in-the-blanks systems, such as Mail-Merge in Microsoft Word.
- *Level 2*: This level is associated with systems that employ scripts or rules to add texts, possibly based on some conditions such as in web templating languages (e.g. Jinja).
- *Level 3*: This is where some linguistic knowledge is utilized to deal with morphology, morphophonology[12] and orthography[13], easing the effort of requiring complex templates.
- *Level 4*: The next stage in sophistication progresses to writing complete sentences and paragraphs from representations of the meaning.
- *Level 5*: The last stage is dynamically creating documents, where the narrative of the whole document is controlled by the system.

# 4 DISCUSSION AND CONCLUSION

## 4.1 KEY FINDINGS

In this section, we will highlight key findings from the reviewed studies. For the purposes of this survey, we have looked at how various domains have modified the basic document generation workflow to suit their requirements and add efficiency to the process. We primarily analyzed the content authoring and document assembly steps. Table 5 provides a summary of the reviewed approaches in terms of applicable documents, common features, distinct characteristics and underlying technologies. Here are key findings:

- One recurring theme in product technical DA architectures is the use of content standards such as DITA and DocBook. DITA better supports modularity and reusability of contents generated for a document to be reused in other documents. DITA better supports single sourcing.
- Architectures designed to generate quantitative analysis reports include code execution capacity in the content authoring interface. Such architectures make an explicit attempt to facilitate integrated analysis and documentation.
- Model-based systems engineering approaches generate reports from pre-designed 'model views', putting the spotlight for content authoring on automatic extraction from a central model data repository.
- As for document assembly, the Document Product Line architecture allows variability selection and assembles the relevant fragments of the document. Document assembly systems based on a rule base and user responses to a questionnaire have also been developed.
- Another key finding is the availability of a variety of standards to represent different document layers including content, structure, layout, formatting and metadata (for example, see the layers defined in the Akoma Ntoso XML standard). The choice of document representation standard is notable when we design a DA system considering the requirements of desirable operations we want to apply on documents. The DA system architects should choose representation standards that better supports reusability, interoperability, performance, as well as operations such as automated auditing, reasoning over content and quality checks.

---

[12] Morphophonology: The study of the phonological representation of morphemes. (Oxford Dictionary)
[13] Orthography: The study of spelling and how letters combine to represent sounds and form words. (ibid.)

*Table 5. Summary and comparison of DA architectures reviewed*

| DA approach | Document variety | Common features | Distinct characteristics | Technologies used | Related studies |
|---|---|---|---|---|---|
| Quantitative analysis reports | Statistical analysis and scientific reports | Integrating and presenting analysis code, code output and natural language explanation | Reproducibility, usability | Jupyter Notebooks, Integration of R/S into other markup languages. | 53, 63, 94, 112, 99, 43, 37, 5, 105 |
| Legal document assembly tools | Legal documents, contracts, etc. | Dynamic questionnaire; Capture and reuse inputs; Ability to pull info from databases; Questionnaire flexibility; Ability to calculate results; Ability to handle conditional logic; Ability to complete PDF forms; Ability to handle inserted templates | Power drafting, template creation and maintainability | Commercial tools: Hotdocs, Contract Express, etc. | 30, 38, 59, 86, 71, 85 |
| Technical documentation | Software/product lifecycle documents | 1. Documents in regulated software industry to explain development and V&V activities: architecture, functionality, and quality attributes. 2. User guides on how the product works. | Traceability, automatic code documentation | DITA, Programming languages | 20 |
| Model driven engineering | V&V activities: requirements, plan and design, test cases, publish results | Integrating document generation with modeling platform/tool. | Traceability | SysML, XML, DocBook, Latex | 14, 9, 75, 7 |
| Clinical applications of Arden syntax | Documents in daily clinical use such as Discharge letters. | Integration with patient data management systems, templating language should easily integrate with medical domain knowledge | Usability, domain-based, easy to learn. | Arden syntax | 50, 51, 52 |
| Public administration | Documents meant to serve a instructions or rules to a large public. | Document assembly is often the focus of such architectures. | Tailored content, variability driven, personalization | Custom software | 73, 8, 88 |
| Data-to-text | Reports based on machine collected and processed data. | No manual authoring. NLG may be involved to convert data to text. | Informative, periodic, natural language generation | NLG | 28, 106, 29, 107 |
| Highly configurable products | Manuals for each configuration of the product. | Assembly of documents according to product configuration is essential. | Reusability, usability, variability identification | DPL, DITA | 95, 32, 88 |
| Semantic web | Usually used where rule bases are common or for personalization or inter-operability. | Usage of ontologies to improve semantic interoperability, knowledge graphs for generating inferences for document assembly. | Semantic web technology is used. | XML, Custom software | 71, 117, 77, 8, 92, 119 |

Moreover, certain emerging research trends were identified as a part of the survey:

- Personalization has been an emerging field of research in document automation. Documents can be personalized in the content authoring part or in the document assembly part. For example, the Document Product Line architecture was employed to deliver personalized tax forms by assembling the document based on the variability points selected by the users. Similarly an ontology context model was built to personalize the textual content of public communications based on user profile.
- Another theme that can be observed in the reviewed studies is the use of semantic web technologies such as RDF and OWL to perform some level of reasoning over the document content for compliance and auditing purposes and also to automatically assess document quality, structure and formatting against certain policies and standards.

## 4.2 JUPYTER NOTEBOOKS FOR DOCUMENT AUTOMATION

From a conventional documentation perspective, the Jupyter ecosystem, including Jupyter notebooks and the open-source utilities developed for notebooks, presents a great potential for its adoption as a documentation system. Due to the programmatic nature of Jupyter notebooks, they offer a variety of advantages in a DA workflow, such as the following:

- Ability to retrieve information from external databases;
- Ability to add more automation features using Python libraries such as 'ipywidgets', which offer graphical user interface control so the users can decide how their documents should be composed or modified, making the documentation process easier;
- Ability to offer advanced features, such as collaboration and authentication, in Jupyter notebooks;
- Support for enterprise scalability provided through JupyterHub;
- Open-source support and active development of Jupyter notebooks and associated utilities;
- Support for a wide variety of programmatic languages that can work with notebook's underlying standard format, i.e. JSON. Working with this format in the notebook-based DA systems is more convenient than working with XML, which is common in many DA systems. The JSON representation makes it easier to configure and modify notebooks, encouraging the development of a host of tools for merging, processing and validating notebooks. Jupyter notebooks successfully achieve all three promises put forward by XML-based architectures:
    - Jupyter notebooks along with templating engines ensure the separation of content from presentation/formatting.
    - Text can be added uniformly in Markdown and converted to the final document format (Latex/PDF/HTML/DOCX) using Pandoc (MacFarlane 2012).
    - The requirement of a specialized markup is obviated by the fact that HTML and Latex can be directly added into 'Markdown cells' of Jupyter notebooks, which are then processed by Pandoc (although with some limitations related to cross-conversion). Any special markup requirement can also be handled programmatically through post-processing the templating engine conversion output or directly modifying the conversion process.

The biggest advantage of the Jupyter ecosystem is its open-source nature and its use of JSON as the notebook document format. These advantages facilitate the ecosystem's constant development.

However, some aspects of Jupyter notebooks present challenges in wide-spread adoption of notebooks purely for documentation purposes:

- The prominent one is providing graphical text editing capabilities.
- The start-up time and effort for Jupyter notebook is way beyond the expectation for a documentation tool.
- Finally reusability needs to be addressed directly, allowing content reuse from other notebooks, and possibly other sources.

A variety of solutions, including Notebook Extensions and JupyterLab, may be utilized to address some of these challenges.

Overall, the Jupyter ecosystem has a promising future in the documentation sphere and we expect to see more features and developments which may facilitate faster and easier documentation workflows.

## 4.3 FUTURE RESEARCH DIRECTIONS

### 4.3.1 AI for document automation

As more innovative machine learning and AI technologies emerge, there are opportunities to enhance functionality, efficiency and usability of DA tools and supporting tasks:

*Document intelligence*
AI/ML for document reading, document understanding, document analysis, information extraction, question answering, document structure analysis, computer vision, natural text understanding. For a discussion on document intelligence research opportunities, see Motahari et al. (2019) at NeurIPS 2019.

*Intelligent document process awareness*
(1) Automatically anticipate and mitigate document workflow exceptions; (2) Automatically identify changes in input data types and schema and make dynamic process changes; (3) Automatically find and fix missing or incorrect information in the document; (4) Automatically update user guides based on any changes in the product.

*Intelligent document processes optimization*
(1) Automatically suggest and make modifications to document processes to improve overall flow; (2) Automatically learn from past usage logs to figure out better ways to handle document workflows and processes; (3) Automatic orchestration of multiple bots to optimize documentation processes.

*Neural NLG*
High quality data-to-text generation utilizing the latest advances in deep neural networks including transformer architectures based on language models (e.g. BERT and GPT2). Dale (2020) in his review of commercial NLG tools notes that existing commercial NLG technologies for data-into-text generation are conceptually very simple, but effective and useful for many business use cases. He envisages a future for NLG technology which is "a completely different beast, which will see its application in quite different contexts. In particular, neural NLG is set to redefine our notion of authorship. The neural NLG trends indicate that considerably high proportion of texts in the future can be co-authored with machine assistance." (Dale 2020)

### 4.3.2 Making technologies user- and developer- friendly
XML/RDF/ontologies are very potent but have not been adopted very quickly as they are hard to learn and work with. One of the reasons for the popularity and quick development of Jupyter notebooks is that it is based on JSON, which is simple and easy to work with.

DA technologies should be UI oriented and end-user friendly. The fruits of XML can reach the masses only when it is behind a very intuitive UI and does not expect them to code in raw XML. Hence more effort is needed to develop tools which not only make the development of document automation workflows easier, but also make it easier for the end-user.

### 4.3.3 Exploring semantic web technology for inter-operability

Establishing an ontology for the document schema used, if not existent already can be a step towards achieving enterprise-wide semantic interoperability. This will entail seamless flow of data from one department to the other through documents, which can be easily parsed and modified. Enterprise knowledge graphs can also serve as a source for direct content for the documents or can be used to generate inferences which affect how the document is assembled.

### 4.3.4 Integration with domain specific information systems

Domain specific information systems and repositories need better integration with standard documentation technologies. This will allow common DA technologies such as XML to easily pull and push data from such repositories eliminating the manual overhead for users. The changes in the central repository of data should be traceable to the resulting change in a relevant document. Dual traceability should also be established, taking inspiration from MDE based documentation systems. Changes in documents should get replicated in the central data repository and should be traceable, along with the conventional route of changes in repository getting replicated in documents. This will allow seamless flow of information between documents and repositories of data.